\documentclass[letterpaper, 10 pt, conference]{ieeeconf}  

\IEEEoverridecommandlockouts                              

\usepackage{algorithm}
\usepackage{algorithmicx}
\usepackage{algpseudocode}
\usepackage{booktabs}
\usepackage{setspace}
\usepackage{subfig}
\usepackage{multirow}
\usepackage{graphicx}
\usepackage{cite}
\usepackage{color}
\usepackage{tabularx}

\title{\LARGE \bf
Preparation of Papers for IEEE Sponsored Conferences \& Symposia*
}

\title{\LARGE \bf Federated Transfer  Reinforcement Learning for Autonomous Driving
}

\author{Xinle Liang$^{1}$, Yang Liu$^{1*}$, Tianjian Chen$^{1}$, Ming Liu$^{2}$  and Qiang Yang$^{1}$
\thanks{$^{1}$Xinle Liang, Yang Liu, Tianjian Chen and Qiang Yang are with Department of Artificial Intelligence, Webank, Shenzhen, China. \{madawcliang, yangliu, tobychen, qiangyang\}@webank.com} 
\thanks{$^{2}$Ming Liu is with the Robotics and MultiPerception Laborotary, Robotics Institute,  Hong Kong University of Science and Technology, Hong Kong SAR, China. eelium@ust.hk}
\thanks{$^{*}$Corresponding author}
}
\begin{document}

\maketitle
\thispagestyle{empty}
\pagestyle{empty}

\begin{abstract}

Reinforcement learning (RL) is widely used in autonomous driving tasks and training RL models typically involves in a multi-step process:  pre-training RL models on simulators, uploading the pre-trained model to real-life robots, and fine-tuning the weight parameters on robot vehicles. This sequential process is extremely time-consuming and more importantly, knowledge from the fine-tuned model stays local and can not be re-used or leveraged collaboratively.   To tackle this problem, we present an online federated RL transfer process for real-time knowledge extraction where all the participant agents make corresponding actions with the knowledge learned by others, even when they are acting in very different environments. To validate the effectiveness of the proposed approach, we constructed a real-life collision avoidance system with Microsoft Airsim simulator and NVIDIA JetsonTX2 car agents, which cooperatively learn from scratch to avoid collisions in indoor environment with obstacle objects.  We demonstrate that with the proposed framework, the simulator car agents can transfer knowledge to the RC cars in real-time, with 27\% increase in the average distance with obstacles and 42\% decrease in the collision counts.

\end{abstract}

\section{INTRODUCTION}

Recent Reinforcement Learning (RL) researches in autonomous robots have achieved significant performance improvement by employing distributed architecture for decentralized agents \cite{chen2017decentralized,desjardins2009learning}, which is termed as Distributed Reinforcement Learning (DRL). However, most existing DRL frameworks consider only synchronous learning with a constant environment. In addition, with the fast development of autonomous driving simulators, it is now common to perform pre-training on simulators, and then transfer the pre-trained model to real-life autonomous cars for fine-tuning. One of the main drawbacks of this path is that the model transfer process is conducted offline, which may be very time-consuming, and there is lack of feedback and collaborations from the fine-tuned model trained with different real-life scenarios. 

To overcome these challenges, we propose an end-to-end training process which leverages federated learning (FL, \cite{Yang2019Federated}) and transfer learning \cite{pan2009survey} to enable asynchronous learning of agents from different environments simultaneously. Specifically, we bridge the pre-training on simulators and real-life fine tuning processes by various agents with asynchronous updating strategies. Our proposed framework alleviates the time-consuming offline model transfer process in autonomous driving simulations while allows heavy load of training data stays local in the autonomous edge vehicles. Therefore the framework can be potentially applied to real-life scenarios where multiple self-driving technology companies collaborate to train more powerful RL tasks by pooling their robotic car resources without revealing raw data information. We perform extensive real-life experiments on a well-known RL application, i.e, steering control RL task for collision avoidance of autonomous driving cars to evaluate the feasibility of the proposed framework and demonstrates that the proposed framework has superior performance compared to the non-federated local training process.
\begin{figure*}[!t]
\centering
\subfloat[JetsonTX2 RC car]{\includegraphics[width=1.1in]{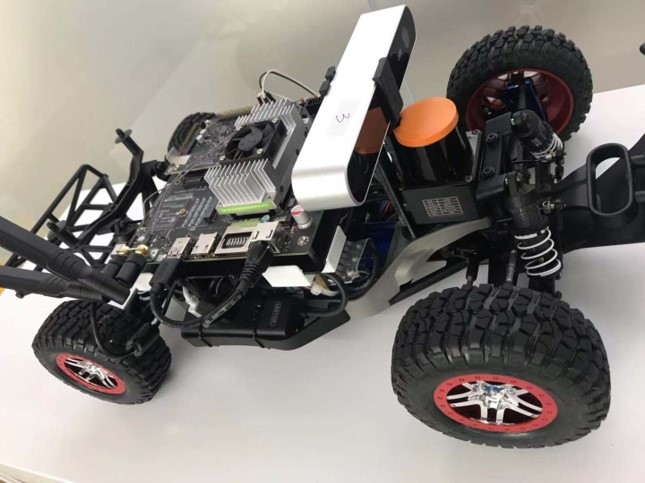}%
\label{jetsoncar}}
\hspace{18mm}
\subfloat[``coastline" map in Airsim platform]{\includegraphics[width=1.1in]{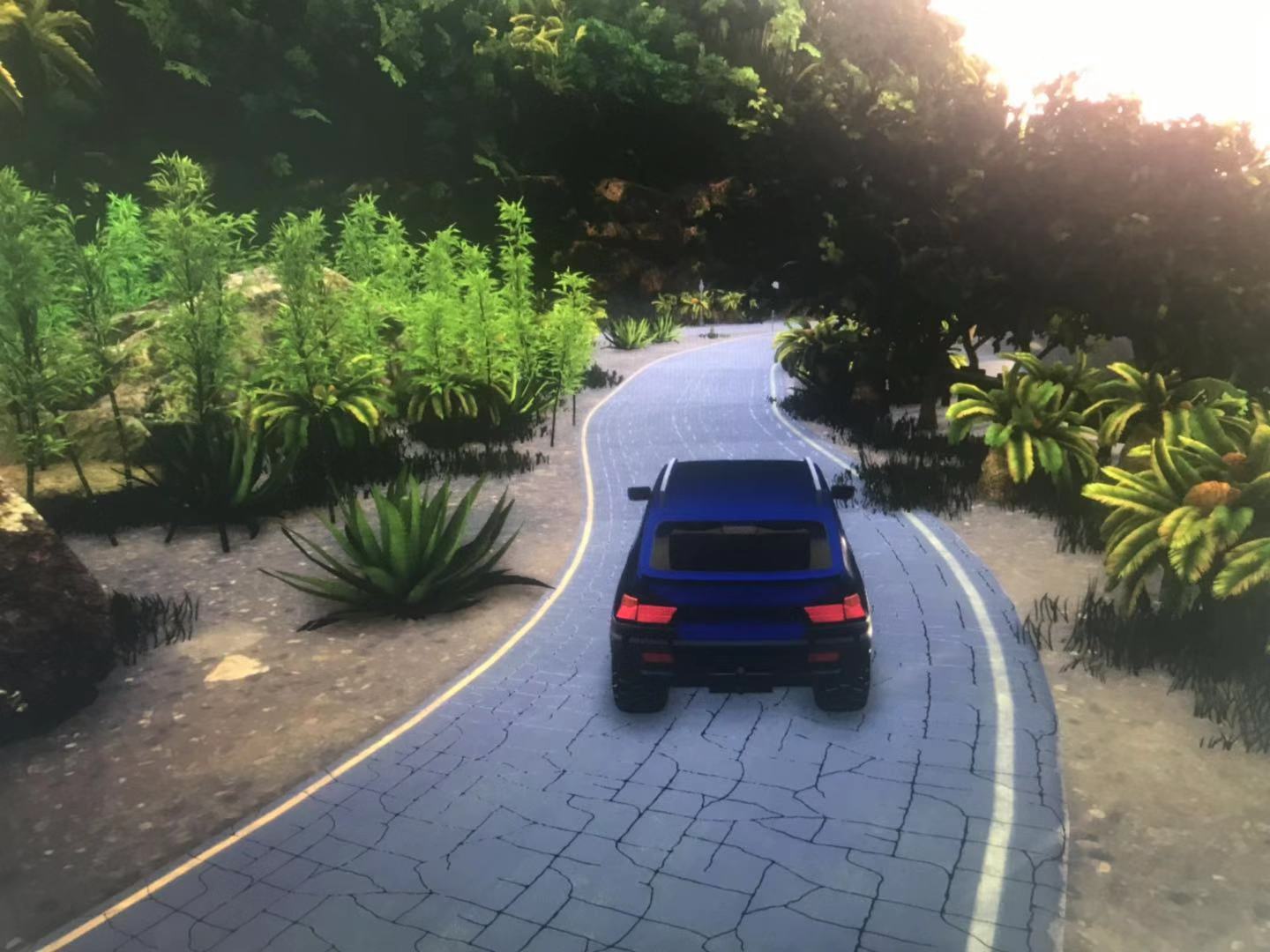}%
\label{airsim}}
\hspace{18mm}
\subfloat[Experiment Race]{\includegraphics[width=1.1in]{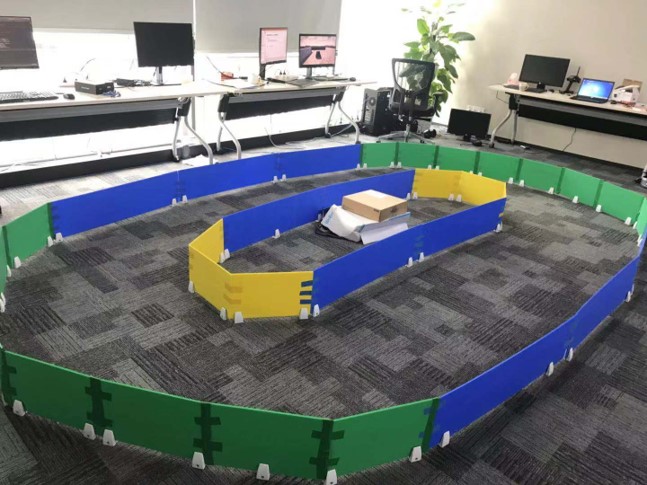}%
\label{track}}
\caption{Hardware and simulator platforms employed in FTRL validation experiments. }
\label{figure_1}
\end{figure*}

\subsection{Related Work}
\label{Ineffectiveness of DRL}

One of the most important tasks for transfer reinforcement learning is to generalize the already-learned knowledge to new tasks  \cite{parisotto2015actor,barreto2017successor,ma2018universal}. With the fast advance of robotics simulators, lots of researches start to investigate the feasibility and effectiveness of transferring the knowledge of simulators  to real-life agents \cite{chen2017decentralized,long2018towards,yuan2019end,pan2017virtual,cutler2016autonomous}.

\cite{long2018towards} proposed a decentralized end-to-end sensor-level collision avoidance policy for multi-robot systems, with the pre-trained process conducted on  stage mobile robot simulator$\footnote{http://rtv.github.io/Stage/}$. \cite{chen2017decentralized} studied the problem of reducing the computationally prohibitive process of anticipating interaction with neighboring agents in a decentralized multi-agent collision avoidance scenario. The pre-trained model of the RL model used is based on the trained data generated by the simulator. \cite{yuan2019end} investigated the problem end-to-end nonprehensile rearrangement that
maps raw pixels as visual input to control actions without any form of engineered feature extraction. The authors firstly trained  a suitable rearrangement policy in Gazebo \cite{koenig2004design}, and then adapt the
learned rearrangement policy to real-world input data based on the transfer framework proposed. 

It can be easily concluded that for transfer reinforcement learning in robotics area, most RL researches employed the following research path: pre-training RL model on simulators, transferring the model to robots and fine-tuning the model parameters.  Usually, the above processes are executed sequentially, i.e., after the RL models have been pre-trained and transferred to the robots, no meaningful experience or knowledge from the simulators can be provided for the final models fine-tuned on the real-life robots. Then, one may ask: can we make the transfer and fine-tune processes  executed in parallel?

The framework proposed in this work utilizes RL tasks in the architecture of  federated learning. Note that some recent works also investigate federated reinforcement learning (FRL) architecture. \cite{Hankz2019} presents two  real-life FRL examples for privacy-preserving issues both in manufactory industry and medical treatment systems. The authors further investigated the problem of multi-agent RL system in a cooperative way, when considering the privacy-preserving requirements of agent data, gradients and models. \cite{Liu2019} studied the FRL settings in the autonomous navigation where the main task is to make the robots fuse and transfer  their experience so that they can effectively use prior knowledge and quickly adapt to new environments. The authors presented the Lifelong Federated Reinforcement Learning (LFRL), in which the robots can learn efficiently in a new environment and extend their experience so that they can use their prior knowledge. \cite{nadiger2019federated} employed the techniques in FRL for personalization of a non-player character, and developed player grouping policy, communication policy and federation policy respectively.

\subsection{Our Proposal}

Different from existing FRL researches, our research motivation originates from  the feasibility of conducting online transfer on the knowledge learned from one RL task to another task, with the aim of both  federated learning and online transfer model.

In this paper, we present  Federated Transfer Reinforcement Learning (FTRL) framework, which is capable of transferring RL agents knowledge in real-time on the foundation of federated learning. To the best of our knowledge, it is the first literature dealing with FRL techniques with online transfer model. Compared to the above existing works, our proposed framework has the advantages of
\begin{enumerate}
\item \textit{Online transfer.} The proposed framework is capable of executing the source and the target RL tasks in simulator or real-life environments with non-identical robots, obstacles, sensors and control systems;
\item \textit{Knowledge aggregation.}  Based on the functionality of federated learning, the proposed framework can conduct knowledge aggregation process in nearly real-time.
\end{enumerate}

We validate the effectiveness of FTRL framework on the real-life collision avoidance systems on  JetsonTX2 remote controlled (RC) cars  and  the Airsim simulators. The experiment results show that FTRL can transfer the knowledge  online, with better training speed and evaluation performance.

\section{HARDWARE PLATFORM AND TASKS}

 In order to better illustrate and validate the framework proposed, we construct  real-life autonomous systems based on three JetsonTX2 RC cars, Microsoft Airsim autonomous driving simulator and a PC server. Fig. 1 presents the basic hardware and software platforms used in the validation process.

The real-life RL agents run on three RC cars, which house a battery, a JetsonTX2 single-board computer, a USB hub, a LIDAR sensor and an on-board Wi-Fi module.  Fig. \ref{jetsoncar} presents an image of the experiment RC car.

In the collision avoidance experiment, we use a PC   as the model pre-training platform and as  the FL server, which is armed with an 8-core 32G Intel i9-9820X CPU, and 4 NVIDIA 2080 Ti GPU.

Developed by Microsoft, Airsim is  a simulator for drones and cars, which  serves as a platform for AI research to experiment with ideas on deep reinforcement learning, autonomous driving etc. 
The version used in this experiment is \textit{v1.2.2.-Windows} \footnote{https://github.com/Microsoft/AirSim/releases}. In the pre-train and federation processes, we ``coastline'' build-in map in the Airsim platform, which can be seen in Fig. \ref{airsim}.

As can be seen in Fig. \ref{track}, we construct a fence-like experimental race for the collision avoidance tasks in indoor environment. We regularly change the overall shape of the race and sometimes set  some obstacles in the race in order to construct different RL environments. However, for a single run of a specific RL task, the race shape and obstacle positions remain unchanged.

\begin{figure*}[htb]
 \center{\includegraphics[width=13cm]  {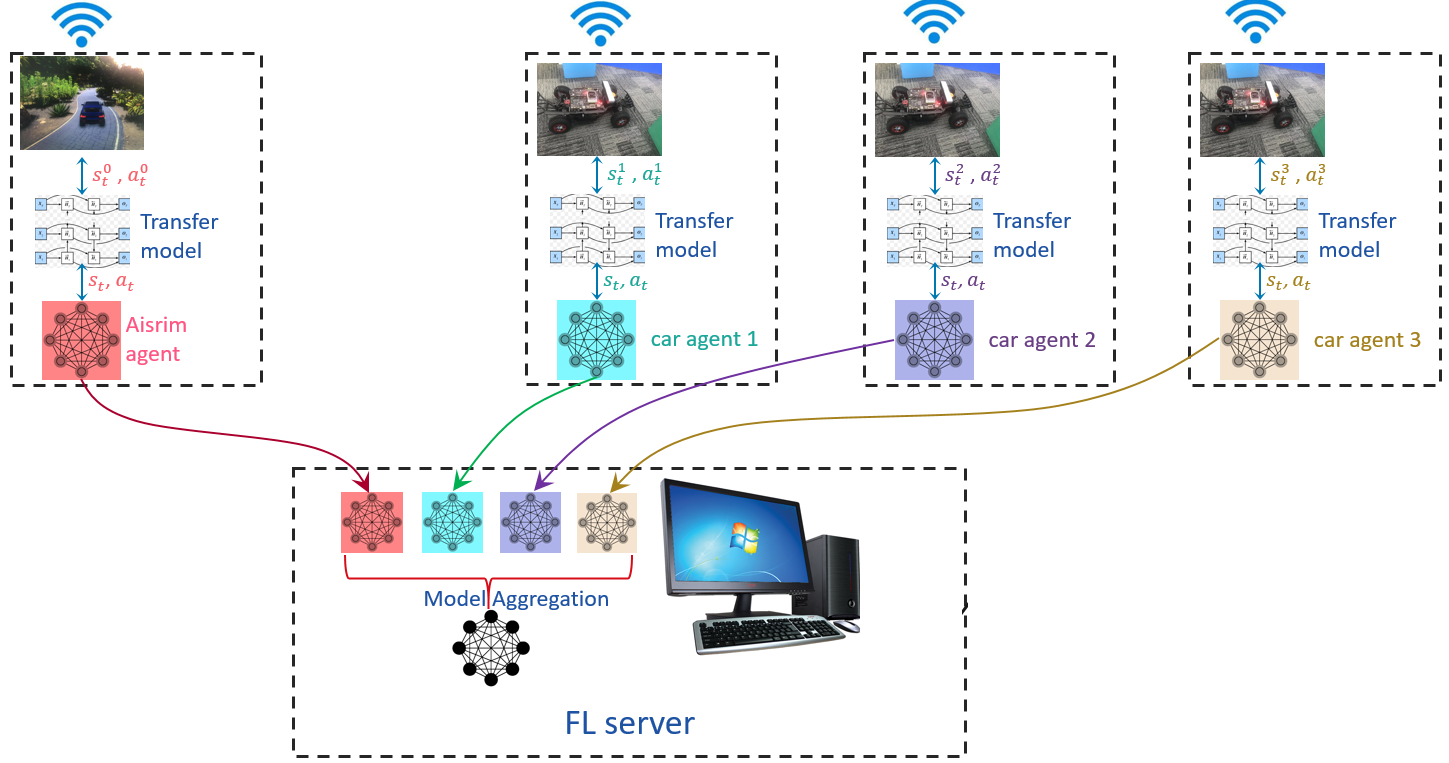}}
 \caption{  The FTRL framework for collision avoidance RL tasks of autonomous driving cars. All participant agents and the FL server communicate through the wireless router. Each agent executes the RL task in its corresponding environment. The FL server regularly aggregates the RL models of all agents and generates the federation model, which is asynchronously updated by different RL agents.}
 \label{FRL_FDQN}
 \end{figure*}

\section{PROPOSED FRAMEWORK}
It is worth noting that FTRL framework is not  designed for any specific RL method. However, in order to thoroughly describe the framework and validate its effectiveness, Deep Deterministic Policy Gradient (DDPG, \cite{Lillicrap2015Continuous}) is chosen to be the RL implementation. 

\subsection{DDPG Algorithm}
We consider the following standard RL setting: A RL agent has to interact with the stochastic environment in discrete time.  At each time step $t$ the agent makes observations $s_t \in \mathcal{S}$, takes actions $a_t \in \mathcal{A}$, and receives rewards $r(s_t,a_t) \in \mathcal{R}$. We  assume that the environments considered in this work have real-valued observations  $\mathcal{S}=\mathcal{R}^d$ and actions $\mathcal{A} = \mathcal{R}^d$. For deterministic action case, the agent's behavior is controlled by a deterministic policy $\mu(s)$: $\mathcal{S} \rightarrow \mathcal{A}$,  which maps
each observation to an action. The state-action value function, which describes the expected return
conditioned on first taking action $a \in \mathcal{A}$ from state $s \in \mathcal{S}$ and subsequently acting according to $\mu$, is defined as
\begin{equation}
Q_{\mu}(s,a) = E[\sum_{t=0}^{\infty} \gamma^t r(s_t,a_t)], 0<\gamma<1
\end{equation}
where $\gamma$  is the discount factor.

DDPG \cite{Lillicrap2015Continuous} is an off-policy actor-critic algorithm, which primarily uses two neural networks, one for the actor and one for the critic. The critic network is updated from the gradients obtained from the temporal difference (TD) error. The actor network is updated by the deterministic policy gradient by Silver et al. \cite{Silver2014Deterministic}.

\subsection{Steering Control RL Settings}
Similar to some existing works on single  collision avoidance system \cite{Llorca2011Autonomous,Soudbakhsh2015Steering}, we seek to develop  steering control to avoid collisions for  autonomous agents. The observations $s_t$ are the LIDAR distance data, which are  collected by the  sensor equipped on the autonomous cars. To accomplish this task, we introduce a specific  reward function conditioned on observations $s_{t+1}$, i.e., $r_t(s_{t+1})$, which is defined as follows:
\begin{equation}
r_t(s_{t+1})=\overline{r}-\overline{c} * cond [min(s_{t+1})<d]-2^{\overline{d}-m_d}
\label{reward_function}
\end{equation}
where $m_d = \frac{1}{\lfloor f*n \rfloor} \sum_{i=0}^{\lfloor f*n \rfloor} \overline{s}_{t+1}$  $n$ is the number of the dimensions of the LIDAR distance data, $f$ is a fraction of the distance data ($0<f<1$), $\lfloor f*n \rfloor$ denotes the maximal integer no larger than $f*n$ and $\overline{s}_{t+1}$ represents the ascending sequence  of  ${s}_{t+1}$, $cond[*]=1$  if event $*$ happens else 0. $\overline{r}$ is a positive base reward value, $\overline{c}$ is a positive penalty value for collision events and $\overline{d}$ is a positive value for casting exponential penalty on $m_d$. It can be concluded from if an action policy is targeted to make good performance, it should obtain: 1) no collision events and 2) to make the smallest $f$ fraction of distance data $m_d$ as great as possible.

Note that we set the reward function $r_t$ to be conditioned on ${s}_{t+1}$ rather than on $s_t$ and $a_t$ based on the following  considerations:
\begin{enumerate}
    \item The collision event caused by $a_t \gets \mu(s_{t})$  can be detected by $s_{t+1}$: when the minimal value of LIDAR data $s_{t+1}$ is lower than the predefined safe distance $d$ i.e., $min(s_{t+1})<d$, then  a collision event is detected, and  thus a  penalty value $\overline{c}$ is activated in the reward function.
    \item  Given the current observation $s_t$, a good steering action policy is capable of making the autonomous agent to stay away from any obstacle in the next state as far as possible. Specifically,  the autonomous agent is expected to maximize its minimal distance with all obstacles in the next time step, i.e., $\mathbf{MAX}{min(s_{t+1})}$.
    Moreover, for the sake of the existence of stochastic factors, we choose to make exponential penalty on the average value of the smallest  $f$ fraction  of the ascending sequence of ${s}_{t+1}$, i.e., $-2^{\overline{d}-m_d}$.
\end{enumerate}

\subsection{FTRL Framework}
For the  collision avoidance task conducted herein, we present the FTRL framework. The basic components of a FTRL framework are presented in Fig. 2. There are different autonomous car agents conducting collision avoidance RL tasks in different environments, including the  real-life  and the simulator environments. All agents share identical model structure,  so that their models can be aggregated by FedAvg process \cite{McMahan2017Communication,Yang2019Federated}. The basic training process is as follows:
\begin{enumerate}
 \item    \textit{Online transfer process.} Since distributed RL agents  are acting in  various environments, a knowledge transfer process is needed when each RL agent interacts with its specific environment;
\item \textit{Single RL agent training and inference.} This process serves as a standard RL agent training and inference process.
\item  \textit{FedAvg process.}  All the useful knowledge of distributed RL agents is aggregated by FedAvg process of the RL models, which can be expressed as:
    \begin{equation}
    w_{fed}^{\theta} \leftarrow \frac{1}{N}\sum_{i=1}^N w_{i}^{\theta}
    \end{equation}
where $w_{fed}^{\theta}$,  $w_{i}^{\theta}$ represent the network parameters of the federation model and the model of the $i$-th RL agent respectively, and $N$ is the number of all RL agents.  $w_{fed}^{\theta}$ is updated element-wisely as the arithmetic mean of all RL models.

\end{enumerate}

\textbf{Online Transfer Process.} Since the RL tasks to be accomplished are highly-relative and all observation data are propositional-correlated and pre-aligned, one possible transfer strategy is to make numeric alignments on the observations and actions.
 According to the reward function  Eq. \ref{reward_function}, $r_t$ is solely dependent on $s_{t+1}$. Therefore, we only have to make transfer process on $s_t$ and $a_t$.
 For the LIDAR observation data, we set one environment as standard environment, and all observations of non-identical scales  can be transformed into the standard observation $s_{t}$ based on the following propositional way:
\begin{equation}
s_{t} = \beta_i s_t^i
\label{observation_model}
\end{equation}
where $\beta_i$ is a super-parameter controlling the scale-ratio of the $i$-th  and the standardized environments. We then   standardize the action of DDPG into  range $(-1,1)$ and when making steering action, the $i$-th agent acts as:
\begin{equation}
\label{action_stand}
    a_t^{i} = a_t {|\mathbf{Max}_{i \in \{1,...,\infty\}}a^i|}
\end{equation}
where $\mathbf{Max}_{i \in \{1,...,\infty\}}a^i$ represents the maximal range of the steering control for a specified car in the $i$-th environment.  The detailed processes for the RL agent and the FL server are presented in the Algorithm 1 and 2 ($\mathcal{N}_i$ presents the DDPG model of the $i$-th agent and $\mathcal{N}_{fed}$ represents the federation model).

\begin{algorithm}
\label{RL_algorithm_i}
        \caption{Training procedure for the $i$-th agent}
        \begin{algorithmic}[1]
            \Require synchronization cycle $t_{u}$, $t_0 \gets $ current time, scale-ratio $\beta_i$
                \While{not terminated}
                \State get current observation $s_t^{i}$
                \If{Transfer process is needed}
                \State $s_t \gets $ TRANSFER\_OBSERVATION($s_t^{i}$)
                \State get $a_t$ from DDPG \cite{Lillicrap2015Continuous}
                \State make steering atcion $a_t^{i} \gets $
                \State TRANSFER\_ACTION($a_t$)
                \EndIf
                \State  Get current time $t_1$
                \If{$t_1 - t_0 > t_{u}$}
                        \State $t_0 \gets t_1$
                        \State UPDATEMODEL( )
                    \EndIf
               \State train local $\mathcal{N}_i$ with DDPG \cite{Lillicrap2015Continuous}
                \EndWhile
            \Function {TRANSFER\_ACTION}{$a_t$}
                \State $a_t^{i} = a_t {|\mathbf{Max}_{i \in \{1,...,\infty\}}a^i|}$
                \State return $a_t^{i}$
            \EndFunction
            \State
            \Function {TRANSFER\_OBSERVATION}{$s_t^i$}
                \State $s_{t} = \beta_i s_t^{i}$
                \State return  $s_{t}$
            \EndFunction
            \State

            \Function {UPDATEMODEL}{ }
                \State get federated model $\mathcal{N}_{fed}$ from FL server
                \For{$w_{fed}^{\theta}$ in $\mathcal{N}_{fed}$}:
                \State $w^{\theta} \leftarrow w_{fed}^{\theta} $
                \EndFor
            \EndFunction
            \State
           
        \end{algorithmic}
    \end{algorithm}

The training procedure for FTRL works in an asynchronous way:
\begin{enumerate}
    \item \textit{The $i$-th agent procedure}. As can be seen in Algorithm 1, for the $i$-th agent, firstly, according to Eq. \ref{observation_model}, an agent-specified transfer process is employed if the current agent is not acting in the standard environment. Then it asynchronously updates the RL model from the  FL server if needed. Lastly, it trains the RL model  from the experience buffer with DDPG algorithm.  A super-parameter $t_u$ is introduced in order to control  the time interval of updating the federation model from the FL server.
    \item \textit{FL server procedure}. As can be seen in Algorithm 2, the FL server regularly collects all the RL models from all agents, which is controlled by the super-parameter federation cycle $t_f$. Then the FL server generates the federation model by FedAvg process.
\end{enumerate}

The inference  for FTRL is rather simple:
 the $i$-th agent receives the observation $s_t^i$ and then, if needed, performs transfer process  according to Eq. \ref{observation_model}. Then the standardization action  can be computed  by  $a_t \gets \mu_i(s_t)  + \mathcal{U}_t$ ($\mathcal{U}_i$ denotes the $t$-th time step result of the random process $\mathcal{U}$ in DDPG),   and lastly the steering action $a_t^{i}$ can be made by  Eq. \ref{action_stand}.

Note that since Algorithm 1 and 2 work asynchronously, some weights update  process of $\mathcal{N}_i$ of local RL agents may not be used. For example, we assume that two model synchronization processes of the $i$-th agent happen at time $t_0^i$ and $t_1^i$ respectively, a federation process of the FL server happens between the two synchronization processes at time  $t_0^{fed}$, i.e., $t_0^i<t_0^{fed}<t_1^i$. Since at time $t_1^i$, this agent updates it model to the federation model generated at time $t_0^{fed}$,  the local training processes between time $t_0^{fed}$ and $t_1^i$ makes no impact to the FL system. It is trivial to extend the current framework to conduct asynchronous model updates, similar to \cite{Liu2019}.

\begin{algorithm}
        \caption{Federation procedure  for FL server}
        \begin{algorithmic}[1]
            \Require federation cycle $t_{f}$, $t_0 \gets$ current time
            \Ensure
                \While{not terminated}
                \State  get current time $t_1$
                \If{$t_1 - t_0 > t_{f}$}
                        \State $t_0 \gets t_1$
                        \For{$i$ in 1...N }
                        \State get single model $\mathcal{N}_i$
                        \EndFor
                        \For{$w^{\theta}$ in $\mathcal{N}$ }
                            \State $w^{\theta}_{fed} \gets \frac{1}{N}\sum_{i=1}^N w^{\theta}_{i}$
                        \EndFor
                \EndIf
                \EndWhile
        \end{algorithmic}
    \end{algorithm}


\section{Experiments}

\begin{figure*}[!t]
\centering
\subfloat[DDPG]{\includegraphics[width=1.8in]{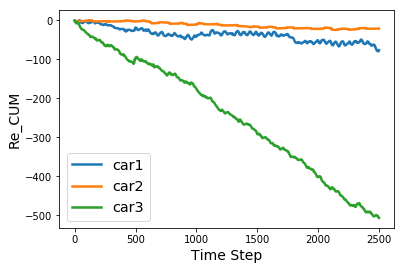}
\label{ddpg_result}}
\hspace{4mm}
\subfloat[FTRL-DDPG]{\includegraphics[width=1.8in]{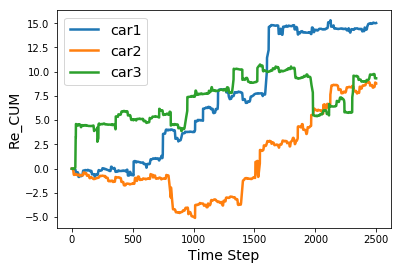}
\label{fed_ddpg_result}}
\hspace{4mm}
\subfloat[FTRL-DDPG-SIM]{\includegraphics[width=1.8in]{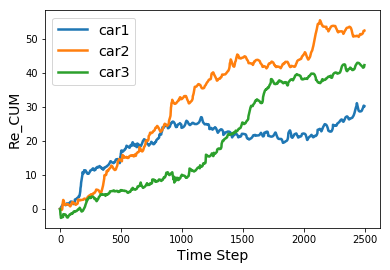}
\label{simu_fed_ddpg_result}}
\caption{Cumulative  summation of \textit{relative performance} for each car with DDPG, FTRL-DDPG and FTRL-DDPG-SIM.}
\label{sole_RL}
\end{figure*}

In this section, we conduct real-life experiments on RC cars and Airsim  in order to validate the followings: 1) FTRL is capable of transferring online knowledge  from simulators to real-life environments; 2) Compared with  a single run of DDPG,  FTRL framework can achieve a better training speed and performance.

\begin{table*}
	\centering
	\begin{tabular}{c|c|cc|cc|cc}
	   \toprule[1.8pt]
		\multicolumn{2}{c|}{\multirow{2}*{}} & \multicolumn{2}{|c|}{\textbf{car1}} & \multicolumn{2}{c|}{\textbf{car2}} & \multicolumn{2}{|c}{\textbf{car3}} \\
		\cline{3-8}
		\multicolumn{2}{c|}{~} & avg\_dist & coll\_no & avg\_dist & coll\_no & avg\_dist & coll\_no \\
\midrule[1.pt]
    \multicolumn{2}{c|}{DDPG} & 0.39 & 18 & 0.29 & 31 & 0.38 & 24 \\
		\multicolumn{2}{c|}{FTRL-DDPG} & 0.42 (7.7\%) & \textbf{9} (50\%) & 0.37 (27.6\%) & 27 (12.9\%) & \textbf{0.51} (34.2\%) & 17 (29.2\%) \\
		\multicolumn{2}{c|}{\textbf{FTRL-DDPG-SIM} }& \textbf{0.45} (15.4\%) & 12 (33.3\%) & \textbf{0.39} (34.5\%) & \textbf{16} (48.4\%) & 0.50 (31.6\%) & \textbf{13} (45.8\%) \\
    \bottomrule[1.8pt]
	\end{tabular}
	\caption{The averge distance and collision number results of three JetsonTX2 RC cars on the test experiments in Fig. 4. For each  approach on each car, 50 cycles in the race are executed. The results in bold denote the better results for each car (with smaller average distance or collusion count). }
	\label{reward_count_table}
\end{table*}

\subsection{Application Details}
In this subsection, for the sake of reproductions, we are going to present the basic application settings for  FTRL, Airsim platform and the  RC cars.

\textbf{(The codes for all the implementations are uploaded to http...... )}. The following presents the basic DDPG settings employed: the actor network is equipped with  three  128-unit fully-connected layers with a continuous output layer, while the critic  network also has three  128-unit fully-connected layers with a state-action output layer; We set $\gamma  \gets 0.99$ and $\tau \gets 0.02$, and learning rates for both actor and critic networks  1e-4. We set the experience buffer size to be 2500 and  batchsize  32.

The basic settings for Airsim  is in the uploaded setting file $settings.json$: in order to maintain a good transferrability to the  RC cars, the LIDAR sensor is set to be only able to collect the distance data of the front view (with `HorizontalFOV' range [-90,90]), which are divided into 60 dimensions from left to right. We use the public build-in map ``coastline" of Airsim to conduct the pre-training  and the federation processes.

In the experiments conducted, we set $\beta_i \gets 6.67$ for all  RC cars. The LIDAR data are collected at a frequency of 40Hz. The interactions among the DDPG agents, the RC car control system and the Airsim are divided into discrete decision making problem with time interval of 0.25 seconds. The federation cycle $t_f$ of the FL server is set to be 2 minutes and the synchronization cycle of local agents $t_u$ 3 minutes.

For the reward function presented in Eq. \ref{reward_function}, we set the base reward value $\overline{r} \gets 8$, the collision penalty value $\overline{c} \gets 60$, the minimum safe distance $d \gets1.1$ and the exponential distance penalty value $\overline{d} \gets 7$.

\subsection{Comparison Results}

Since training DDPG algorithm from scratch on real-life autonomous  cars may take unacceptable time, we have pre-trained a common DDPG model on Airsim platform for all participant DDPG agents. With the pre-trained model, each  car can make reasonable action corresponding to the LIDAR data, which however still has room for improvement. 

In this fine-tune processes of any real-life  agent, we divide the training time  of each DDPG agent into three stages,  with each containing 2500 discrete time steps. Since only the inference of the pre-trained model happens when the number of the experience buffer is smaller than 2500, we ignore the results of the first stages and name the following two stages as stage $I$ and stage $II$. As mentioned, each time step takes 0.25 seconds, and  stage $I$ and  stage $II$ have range $[625,1250)$ and $[1250,1875)$ seconds, respectively.

Since all cars may be running in different environments,  the rewards may be of non-identical scales.  In order to make the results comparable,  the following presents the metric  \textit{relative performance} employed. For a corresponding index $i$ ($1\leq i \leq 2500$)   in stages $I$ and $II$, let $r^{I}_i$, $r^{II}_i$ represent the respective rewards, and the \textit{relative performance} is defined as:
\begin{equation}
    rp_i = \frac{r^{II}_i - r^{I}_i}{r_{max}-r_{min} }
    \label{relative_performance}
\end{equation}
where $r_{max}$ and $r_{min}$  denote  the maximal  and the minimal reward  values for a single run of each car, respectively. It can be  concluded from Eq. \ref{relative_performance} that $-1 \leq rp_i \leq 1$ and  $rp_i>0$ indicates that for the corresponding $i$, the $i$-th time step in stage $II$ performs better than that in stage $I$.

We  keep track of the  cumulative  summation of \textit{relative performance} for different stages, and present the results in  Fig. \ref{sole_RL} of different application  settings, including
\begin{enumerate}
    \item DDPG results on  single RC cars;
    \item FTRL-DDPG results  with the federation of three RC cars(FTRL-DDPG);
    \item FTRL-DDPG results  with the federation of three RC cars and Airsim platform(FTRL-DDPG-SIM);
\end{enumerate}

As can be seen from  Fig. \ref{ddpg_result},  for each car, we can  see that most of the values of the cumulative  summations  of \textit{relative performance} are lower than 0. Moreover, it can be confidently concluded that  the performance decays from stage $I$ to stage $II$. The result indicates that for each run of DDPG, with only 2500 time steps for training, we can make no guarantee on the performance improvements of local RL agents.

However, referring to Fig. \ref{fed_ddpg_result}, for  FTRL-DDPG,  most of the cumulative  summation values of \textit{relative performance} on car1 and car3 are above 0. For car 2, for the first 1500 time steps, an opposite conclusion can be drawn that the performance decays from stage $I$ to stage $II$, and however, for time steps 1500-2500, a significant improvement of the  \textit{relative performance}  can be viewed.  The above results indicate that with FRL framework can accelerate the training speed and improve the performance of the federation of three cars.

Referring to Fig. \ref{simu_fed_ddpg_result}, for  FTRL-DDPG-SIM,  most of the cumulative  summation values of \textit{relative performance} on all cars in the experiments. By comparing the results of FTRL-DDPG-SIM and FTRL-DDPG, we can easily see that FTRL-DDPG-SIM can achieve greater \textit{relative performance} than FTRL-DDPG on most time steps recorded.  The above results indicate that  the transfer model employed in FTRL-DDPG-SIM is effective in accelerating the training speed of autonomous cars by online transferring the knowledge learned from the Airsim simulator, which can take charge of more workload on the training processes of RL agents.

\begin{figure}[!t]
\centering
\includegraphics[width=1.6in]{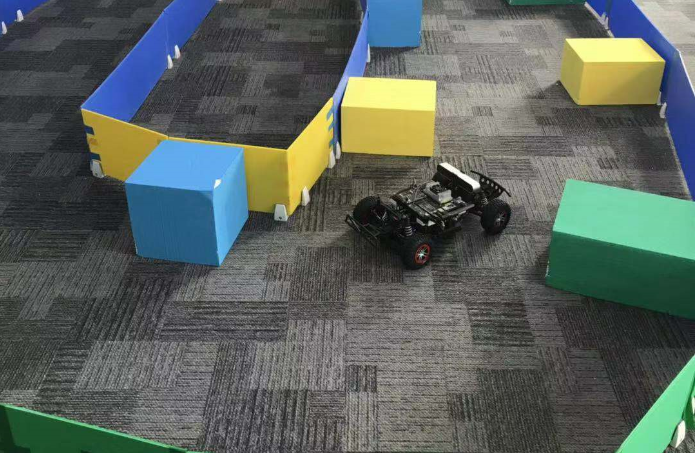}
\caption{The  experimental race with more obstacles for the comparison of the trained model of DDPG and FTRL-DDPG. }
\label{experimental_race}
\end{figure}

In order to better compare all the results of different RL tasks, we further made comparisons on the trained models. The experimental race is shown in Fig. \ref{experimental_race}. It is worth noting that the test race is specifically set to be much more complicated than the  training environments (as shown in Fig. \ref{track}), which is with more obstacles and tighter distances.

For the test experiments with trained models, each run of different cars is executed for 50 cycles in the experimental race. We recorded the average LIDAR distances and the collision numbers for each run of DDPG, FTRL-DDPG and FTRL-DDPG-SIM on each car. It can be easily drawn that a better policy is capable of fulfilling collision avoidance tasks with greater average distance and less collision number. Table \ref{reward_count_table} presents the corresponding results.

As can be seen from Table \ref{reward_count_table}, the results in bold denote the better result for each car. It can be easily seen that for each car, the average distance and collision numbers of FTRL-DDPG  and FTRL-DDPG-SIM are much  less than the corresponding results of DDPG, which demonstrate the effectiveness of FTRL-DDPG-SIM. The following presents an averaging result: for the test experimental race tasks,  compared with DDPG, FTRL-DDPG can make performance improvements with averaging 20.3\% increase in the average distance with obstacles and averaging 30.7\% decrease in the collision number, while for FTRL-DDPG-SIM, the corresponding results are 27.2\% and 42.5\%, respectively.
 
As a conclusion, for the autonomous driving areas, with the capabilities of transferring online knowledge from simulators to real-life cars, FTRL-DDPG-SIM performs better than both single execution of single RL agents and federation model with identical RL agents with better training speed and performance.

\section{Conclusions and Future Work}

In this work, we present the FTRL framework, which is capable of conducting online transfer to the knowledge of different RL tasks executed in non-identical environments. However, the transfer model employed in FTRL presented in this work is rather simple, which is based on human knowledge. Autonomously transferring the experience or knowledge from the already learned tasks to new ones online constitutes another research frontier.

\bibliographystyle{IEEEtrans}

\bibliography{ref}

\begin{thebibliography}{10}
\providecommand{\url}[1]{#1}
\csname url@samestyle\endcsname
\providecommand{\newblock}{\relax}
\providecommand{\bibinfo}[2]{#2}
\providecommand{\BIBentrySTDinterwordspacing}{\spaceskip=0pt\relax}
\providecommand{\BIBentryALTinterwordstretchfactor}{4}
\providecommand{\BIBentryALTinterwordspacing}{\spaceskip=\fontdimen2\font plus
\BIBentryALTinterwordstretchfactor\fontdimen3\font minus
  \fontdimen4\font\relax}
\providecommand{\BIBforeignlanguage}[2]{{%
\expandafter\ifx\csname l@#1\endcsname\relax
\typeout{** WARNING: IEEEtran.bst: No hyphenation pattern has been}%
\typeout{** loaded for the language `#1'. Using the pattern for}%
\typeout{** the default language instead.}%
\else
\language=\csname l@#1\endcsname
\fi
#2}}
\providecommand{\BIBdecl}{\relax}
\BIBdecl

\bibitem{chen2017decentralized}
Y.~F. Chen, M.~Liu, M.~Everett, and J.~P. How, ``Decentralized
  non-communicating multiagent collision avoidance with deep reinforcement
  learning,'' in \emph{2017 IEEE international conference on robotics and
  automation (ICRA)}.\hskip 1em plus 0.5em minus 0.4em\relax IEEE, 2017, pp.
  285--292.

\bibitem{desjardins2009learning}
C.~Desjardins, J.~Laum{\^o}nier, and B.~Chaib-draa, ``Learning agents for
  collaborative driving,'' in \emph{Multi-Agent Systems for Traffic and
  Transportation Engineering}.\hskip 1em plus 0.5em minus 0.4em\relax IGI
  Global, 2009, pp. 240--260.

\bibitem{Yang2019Federated}
Q.~Yang, Y.~Liu, T.~Chen, and Y.~Tong, ``Federated machine learning: Concept
  and applications,'' \emph{ACM Transactions on Intelligent Systems and
  Technology (TIST)}, vol.~10, no.~2, p.~12, 2019.

\bibitem{pan2009survey}
S.~J. Pan and Q.~Yang, ``A survey on transfer learning,'' \emph{IEEE
  Transactions on knowledge and data engineering}, vol.~22, no.~10, pp.
  1345--1359, 2009.

\bibitem{parisotto2015actor}
E.~Parisotto, J.~L. Ba, and R.~Salakhutdinov, ``Actor-mimic: Deep multitask and
  transfer reinforcement learning,'' \emph{arXiv preprint arXiv:1511.06342},
  2015.

\bibitem{barreto2017successor}
A.~Barreto, W.~Dabney, R.~Munos, J.~J. Hunt, T.~Schaul, H.~P. van Hasselt, and
  D.~Silver, ``Successor features for transfer in reinforcement learning,'' in
  \emph{Advances in neural information processing systems}, 2017, pp.
  4055--4065.

\bibitem{ma2018universal}
C.~Ma, J.~Wen, and Y.~Bengio, ``Universal successor representations for
  transfer reinforcement learning,'' \emph{arXiv preprint arXiv:1804.03758},
  2018.

\bibitem{long2018towards}
P.~Long, T.~Fanl, X.~Liao, W.~Liu, H.~Zhang, and J.~Pan, ``Towards optimally
  decentralized multi-robot collision avoidance via deep reinforcement
  learning,'' in \emph{2018 IEEE International Conference on Robotics and
  Automation (ICRA)}.\hskip 1em plus 0.5em minus 0.4em\relax IEEE, 2018, pp.
  6252--6259.

\bibitem{yuan2019end}
W.~Yuan, K.~Hang, D.~Kragic, M.~Y. Wang, and J.~A. Stork, ``End-to-end
  nonprehensile rearrangement with deep reinforcement learning and
  simulation-to-reality transfer,'' \emph{Robotics and Autonomous Systems},
  vol. 119, pp. 119--134, 2019.

\bibitem{pan2017virtual}
X.~Pan, Y.~You, Z.~Wang, and C.~Lu, ``Virtual to real reinforcement learning
  for autonomous driving,'' \emph{arXiv preprint arXiv:1704.03952}, 2017.

\bibitem{cutler2016autonomous}
M.~Cutler and J.~P. How, ``Autonomous drifting using simulation-aided
  reinforcement learning,'' in \emph{2016 IEEE International Conference on
  Robotics and Automation (ICRA)}.\hskip 1em plus 0.5em minus 0.4em\relax IEEE,
  2016, pp. 5442--5448.

\bibitem{koenig2004design}
N.~Koenig and A.~Howard, ``Design and use paradigms for gazebo, an open-source
  multi-robot simulator,'' in \emph{2004 IEEE/RSJ International Conference on
  Intelligent Robots and Systems (IROS)(IEEE Cat. No. 04CH37566)},
  vol.~3.\hskip 1em plus 0.5em minus 0.4em\relax IEEE, 2004, pp. 2149--2154.

\bibitem{Hankz2019}
W.~F. Hankz H.~Z, ``Parallel reinforcement learning,'' in \emph{The 6th World
  Conference on Systemics, Cybernetics, and Informatics}.\hskip 1em plus 0.5em
  minus 0.4em\relax Citeseer, 2019.

\bibitem{Liu2019}
B.~Liu, L.~Wang, M.~Liu, and C.~Xu, ``Lifelong federated reinforcement
  learning: a learning architecture for navigation in cloud robotic systems,''
  \emph{arXiv preprint arXiv:1901.06455}, 2019.

\bibitem{nadiger2019federated}
C.~Nadiger, A.~Kumar, and S.~Abdelhak, ``Federated reinforcement learning for
  fast personalization,'' in \emph{2019 IEEE Second International Conference on
  Artificial Intelligence and Knowledge Engineering (AIKE)}.\hskip 1em plus
  0.5em minus 0.4em\relax IEEE, 2019, pp. 123--127.

\bibitem{Lillicrap2015Continuous}
T.~P. Lillicrap, J.~J. Hunt, A.~Pritzel, N.~Heess, T.~Erez, Y.~Tassa,
  D.~Silver, and D.~Wierstra, ``Continuous control with deep reinforcement
  learning,'' \emph{arXiv preprint arXiv:1509.02971}, 2015.

\bibitem{Silver2014Deterministic}
D.~Silver, G.~Lever, N.~Heess, T.~Degris, D.~Wierstra, and M.~Riedmiller,
  ``Deterministic policy gradient algorithms,'' 2014.

\bibitem{Llorca2011Autonomous}
D.~F. Llorca, V.~Milan{\'e}s, I.~P. Alonso, M.~Gavil{\'a}n, I.~G. Daza,
  J.~P{\'e}rez, and M.~{\'A}. Sotelo, ``Autonomous pedestrian collision
  avoidance using a fuzzy steering controller,'' \emph{IEEE Transactions on
  Intelligent Transportation Systems}, vol.~12, no.~2, pp. 390--401, 2011.

\bibitem{Soudbakhsh2015Steering}
D.~Soudbakhsh and A.~Eskandarian, ``Steering control collision avoidance system
  and verification through subject study,'' \emph{IET intelligent transport
  systems}, vol.~9, no.~10, pp. 907--915, 2015.

\bibitem{McMahan2017Communication}
H.~B. McMahan, E.~Moore, D.~Ramage, S.~Hampson \emph{et~al.},
  ``Communication-efficient learning of deep networks from decentralized
  data,'' \emph{arXiv preprint arXiv:1602.05629}, 2016.

\end{thebibliography}


\end{document}